\documentclass[11pt]{article}

\usepackage[letterpaper,margin=1in]{geometry}
\usepackage[T1]{fontenc}
\usepackage[utf8]{inputenc}
\usepackage{lmodern}
\usepackage{microtype}
\usepackage{amsmath,amssymb,amsfonts}
\usepackage{graphicx}
\usepackage{textcomp}
\usepackage{url}
\usepackage{booktabs}
\usepackage{diagbox}
\usepackage{multirow}
\usepackage{makecell}
\usepackage{tikz}
\usepackage{subcaption}
\usepackage[numbers,sort&compress]{natbib}
\usepackage{hyperref}

\usetikzlibrary{calc,positioning,fit}
\usetikzlibrary{
    arrows.meta,
    positioning,
    calc,
    fit,
    backgrounds,
    decorations.pathreplacing
}

\newcommand{\IEEEPARstart}[2]{#1#2}
\newcommand{\alttext}[1]{}
\newenvironment{keywords}
    {\par\medskip\noindent\textbf{Keywords:}\ }
    {\par}

\hypersetup{
    pdftitle={RECOVAR: Representation Covariances on Deep Latent Spaces for Seismic Event Detection},
    pdfauthor={Onur Efe, Ege Adiguzel, Ali Ozgun Konca, Semih Ergintav, Arkadas Ozakin},
    colorlinks=true,
    linkcolor=blue,
    citecolor=blue,
    urlcolor=blue
}

\title{RECOVAR: Representation Covariances on Deep Latent Spaces for Seismic Event Detection}

\author{
Onur Efe$^{1}$, Ege Adig\"uzel$^{1}$, Ali \"Ozg\"un Konca$^{2}$\\
Semih Ergintav$^{3}$, and Arkadas Ozakin$^{1}$\\[0.4em]
\small $^{1}$Department of Physics, Bo\u{g}azi\c{c}i University, 34342 Istanbul, T\"urkiye\\
\small $^{2}$Department of Geophysics, Bo\u{g}azi\c{c}i University, 34342 Istanbul, T\"urkiye\\
\small $^{3}$Department of Geodesy, Kandilli Observatory and Earthquake Research Institute,\\
\small Bo\u{g}azi\c{c}i University, 34684 Istanbul, T\"urkiye\\
\small Corresponding author: \href{mailto:arkadas.ozakin@bogazici.edu.tr}{arkadas.ozakin@bogazici.edu.tr}
\thanks{This work was supported in part by the Scientific and Technological Research Council of T\"urkiye (T\"UB\.ITAK) under Grant 118C203 (BIDEB 2232 Program), and in part by the TRANSFORM$^2$ Project under Horizon Europe (HORIZON-INFRA-2024-DEV-01-01) under Grant Agreement 101188365.}
}

\date{}

\begin{document}

\maketitle

\begin{abstract}
Modern deep learning methods have improved earthquake detection, but the most widely used approaches typically rely on supervised learning from large labeled waveform datasets. Constructing such datasets is time-consuming and introduces biases from existing catalogs, detection pipelines, and human review, which can limit cross-dataset generalization and sensitivity to underrepresented events. We present RECOVAR, an unsupervised method for seismic event detection that can be trained directly from raw waveforms without using event/noise labels. The method uses an ensemble of convolutional autoencoders and detects events by measuring temporal alignment in their compressed latent representations through a cross-covariance-based metric. The results suggest that RECOVAR can reach the performance of supervised detectors on public benchmark datasets, remain robust under cross-dataset scenarios, and stay effective on imbalanced continuous waveform data where event-containing windows make up less than $1\%$ of the training stream. The method can also complement supervised pipelines by acting as a filter improving precision in a multi-station setting. Because the detection principle makes no domain-specific assumptions, the approach may extend to signal-detection problems beyond seismology.
\end{abstract}


\section{Introduction}
\label{sec:introduction}

\IEEEPARstart{S}{eismic} event detection is a prerequisite for a wide range of seismological analyses, including earthquake cataloging, magnitude estimation, source characterization, and seismic tomography studies. Classical detection algorithms such as short-term average/long-term average (STA/LTA) and waveform-correlation techniques have historically been used to distinguish seismic events from ambient noise, often followed by analyst review~\cite{Allen1978,Withers1998,Gibbons2006}. More recently, supervised deep learning methods have become widely adopted and have improved detection and picking performance across benchmark and regional datasets~\cite{sup:mousavi2020earthquake, sup:soto2021deepphasepick, sup:woollam2019convolutional, sup:zhu2019phasenet, sup:ross2018generalized}.

Despite their benchmark performance, supervised models can exhibit performance variations across different datasets. A review by M\"unchmeyer \emph{et al.} \cite{rev:munchmeyer2022picker} reports performance degradation for several prominent models when they are evaluated across different training and testing domains. Such variation suggests that supervised detectors may partly adapt to the tectonic, instrumental, preprocessing, and labeling characteristics of their training datasets, rather than learning transferable event representations.

The reliance on supervised labels can also restrict discovery in observational seismology. Labeled training datasets are necessarily shaped by existing catalogs, human review, and the classical or supervised detection pipelines used to construct them. As a result, waveform patterns that are absent from existing catalogs or originate from rare or previously unrecognized event types may be treated as noise during training. A high-capacity supervised model trained on such standardized labeled datasets may therefore learn to suppress the signals that are most interesting for discovery.

Unsupervised learning offers a natural route around this limitation. Because it does not require predefined event labels, it can reduce dependence on existing catalogs and has the potential to improve sensitivity to atypical events. However, purely unsupervised seismic event detection remains challenging. Existing applications often rely on clustering followed by manual labeling; thus, a fully automated unsupervised detector has not yet been established.

We propose RECOVAR (REpresentation COVARiances), an unsupervised representation-learning method for seismic event detection. RECOVAR is motivated by the observation that event waveforms originate from a common seismic source and contain coherent temporal patterns that distinguish them from ambient noise. RECOVAR uses an ensemble of independently initialized deep convolutional autoencoders to extract multiple latent representations of the same waveform, and quantifies their shared temporal structure by computing the cross-covariance between them.

\begin{figure}[!ht]
    \centering
    \begin{subfigure}[t]{0.545\textwidth}
        \caption{}
        \centering
        \includegraphics[width=\linewidth]{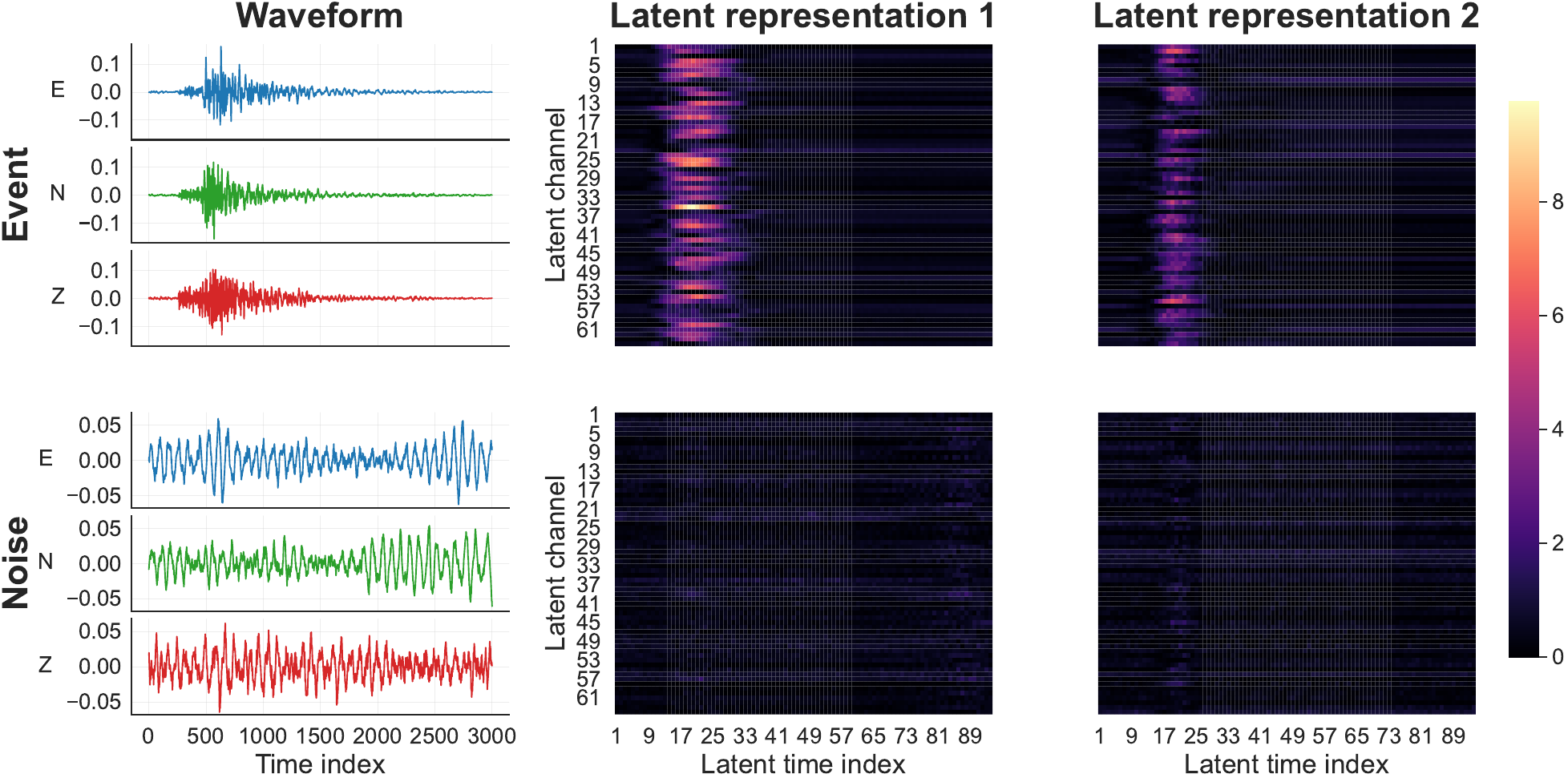}
        \label{fig:raw-latent-examples-event-noise}
    \end{subfigure}
    \hfill
    \begin{subfigure}[t]{0.44\textwidth}
        \caption{}
        \centering
        \includegraphics[width=\linewidth]{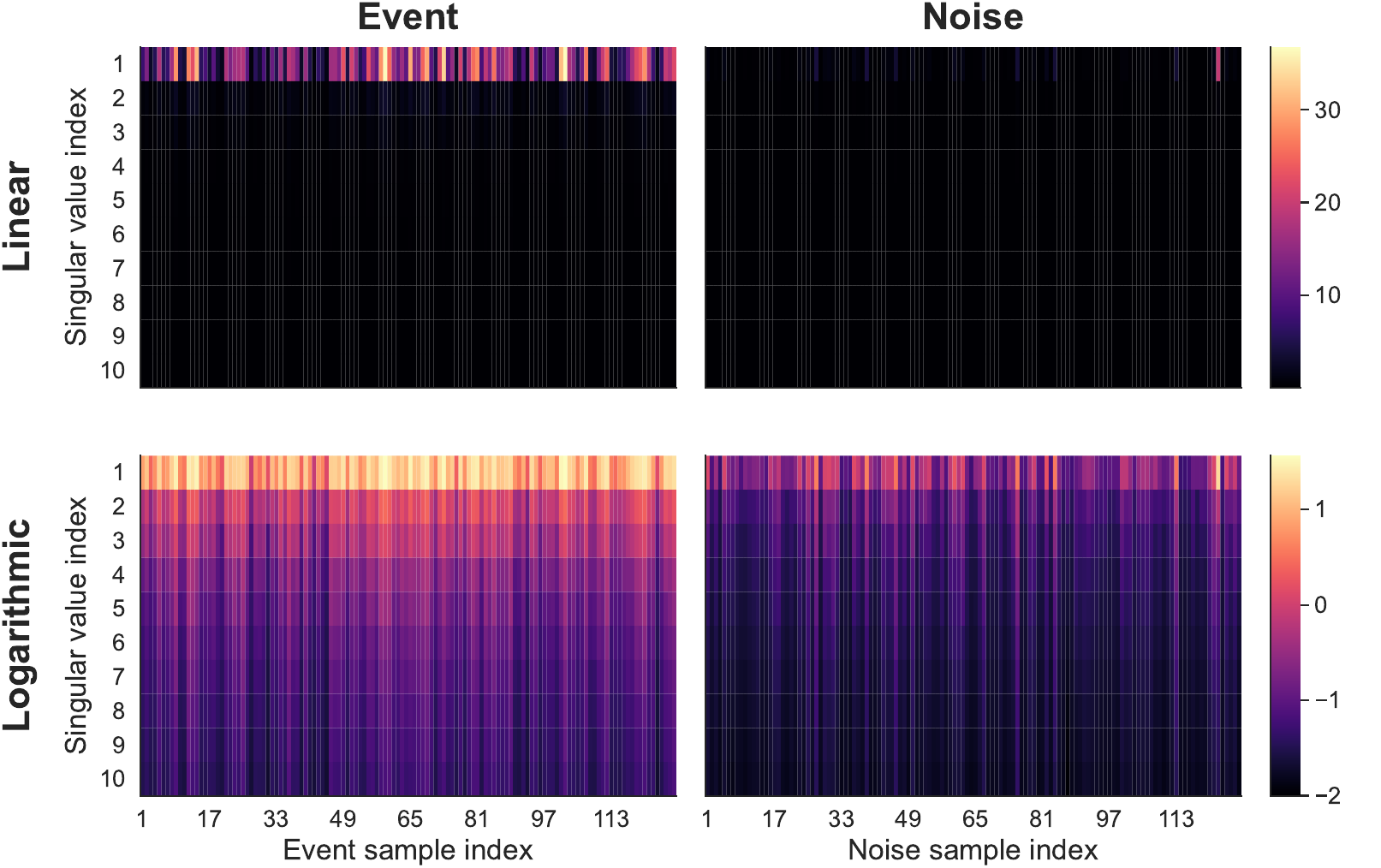}
        \label{fig:singular-spectra-event-noise}
    \end{subfigure}

    \caption{Latent representations of event and noise waveforms exhibit distinct structures. Panel (a) shows example waveform samples and two raw latent representations obtained from different autoencoders. Event representations from different autoencoders show temporally aligned activations, whereas noise representations lack coherent temporal structure and show weaker activations. Panel (b) shows the singular-value spectra of temporal cross-covariance matrices for event and noise samples. For each waveform sample, we compute temporal cross-covariance matrices between all pairs of latent representations and summarize each matrix by its singular values, sorted in descending order and averaged across all autoencoder pairs. Each column is one waveform sample and each row a singular-value component, shown on linear and logarithmic scales. Event samples have a dominant leading singular value, the signature of a common mode of temporal variation across the learned representations. Noise samples have weaker, less concentrated spectra.}
    \label{fig:cross-covariance-and-latent-examples}
\end{figure}

This approach produces a discriminative classification metric, as event waveforms give latent representations with higher covariances across the ensemble than noise waveforms (see Fig.~\ref{fig:cross-covariance-and-latent-examples}). Our experiments replicating the benchmark setting of M\"unchmeyer \emph{et al.} \cite{rev:munchmeyer2022picker} show that the method gives a stable unsupervised detector with performance approaching that of its supervised counterparts and shows smaller performance variation under cross-domain evaluation.

We also propose a training mechanism, which we refer to as \emph{batch dilation}, for learning from continuous seismic recordings where the fraction of event-containing windows is small relative to noise windows ($<1\%$). This mechanism allows the method to maintain detection performance when the model is trained on an imbalanced dataset, and enables applications such as training specialized detectors directly from local continuous waveforms.

Finally, we use RECOVAR to filter detections produced by PhaseNet~\cite{sup:zhu2019phasenet}. In this setting, RECOVAR improves precision under a multi-station coincidence criterion. This result indicates that RECOVAR captures characteristics complementary to those used by the supervised picking and association pipeline, allowing it to reduce spurious detections while preserving most catalog-matched events.
\section{Related Work}\label{related-work}

During the last decade, supervised deep learning methods have opened the door to significant advances in seismic event detection and phase picking. A range of neural-network-based detectors and pickers have been shown to outperform classical triggering algorithms on labeled waveform benchmarks and regional datasets, particularly for low-magnitude or low-SNR events~\cite{sup:mousavi2020earthquake,sup:ross2018generalized,sup:woollam2019convolutional,sup:zhu2019phasenet,sup:soto2021deepphasepick, sup:liu2024seislm}. At the same time, supervised methods depend on large labeled datasets, whose construction often relies on existing catalogs, detection pipelines, and human review. Their performance can also vary under cross-dataset evaluation, suggesting sensitivity to the tectonic, instrumental, preprocessing, and labeling characteristics of the training data~\cite{rev:munchmeyer2022picker}. RECOVAR is motivated by this limitation: it aims to use deep representation learning for event detection without requiring event/noise labels during training.

Unsupervised learning in seismology has most often been used for clustering and exploratory waveform classification. A common strategy consists of representing each waveform by a set of hand-crafted features, applying a clustering algorithm, and then inspecting the resulting clusters, often using a labeled subset or expert interpretation, to determine whether they correspond to meaningful signal classes, such as events, noise, or different event types. For example, Chen \cite{clustman:chen2018fast}, \cite{clustman:chen2020automatic} splits waveforms into segments and applies k-means clustering to manually engineered features extracted from these segments, then manually inspects the resulting classes to use cluster labels as class labels for microseismic events. Duque \emph{et al.} \cite{clustman:duque2020exploring} compare several clustering algorithms, including k-means, BFR, CURE, BIRCH, spectral clustering, and expectation maximization, for classifying seismic events associated with the Cotopaxi volcano. Related approaches use hierarchical clustering or k-means on feature-engineered representations~\cite{clustman:huang2019seismic}, spectral characteristics for separating noise signals~\cite{clustman:johnson2020identifying}, and self-organizing maps applied to Fourier spectra or manually designed feature vectors~\cite{som:carniel2013analysis,som:kuyuk2011unsupervised,som:kohler2010unsupervised}.

More recent work has combined unsupervised learning with learned waveform representations. Mousavi \emph{et al.} \cite{clustnn:mousavi2019unsupervised} use a convolutional autoencoder to obtain feature vectors and then apply k-means clustering to discriminate local events from teleseismic events. Seydoux \emph{et al.} \cite{clustnn:seydoux2020clustering} use a deep scattering network to extract representations, followed by Gaussian mixture clustering to classify earthquake and noise signals. These methods reduce the need for feature design and show that learned representations can capture meaningful waveform structure.

RECOVAR follows a similar representation-learning direction, but differs from previous clustering-based approaches in its objective and evaluation setting. Rather than learning waveform representations and then assigning class labels to clusters via inspection, RECOVAR directly constructs an event-detection score from the shared temporal structure of latent representations learned by an ensemble of autoencoders. This makes the method closer to an unsupervised signal/noise detector than to an exploratory clustering/labeling pipeline. Whereas many previous unsupervised applications have been evaluated in specific case studies, our evaluation emphasizes benchmark event/noise classification, cross-dataset generalization, and training directly from highly imbalanced continuous waveform data.

\section{Method}

\label{sec:method}
\subsection{Overview and Notation}

Our method is based on the observation that event waveforms contain coherent temporal patterns that distinguish them from noise. Motivated by this observation, we train an ensemble of independently initialized convolutional autoencoders to extract latent representations of waveform windows. For event waveforms, these representations are expected to capture coherent temporal patterns, whereas noise is expected to produce weaker and less organized latent activations. As illustrated in Fig.~\ref{fig:cross-covariance-and-latent-examples}, this difference manifests as shared modes of temporal variation across independently learned latent representations. 

To extract these shared modes efficiently, we apply a pointwise linear map to each latent representation. Each convolutional autoencoder is associated with a linear map represented by a trainable square matrix, whose dimension equals the number of channels in the latent space. The same map is applied independently at every latent time step. To encourage the transformed representations to emphasize shared modes of variation, we introduce an additional loss term that promotes temporal alignment. Consequently, transformed representations of event waveform windows exhibit high cross-covariances, which we aggregate to construct an event-detection metric.

The remainder of this section formalizes the components of this framework. Section~\ref{sec:architecture} describes the autoencoder architecture. Section~\ref{sec:training_objective} presents the training objective, including the reconstruction and alignment losses. Section~\ref{sec:detection_metric} defines the quantity which we use as an event detection metric, and Section~\ref{sec:training_procedure} describes the training procedures used for balanced and imbalanced data regimes.

We use the following notation. Scalars are denoted by plain italic letters, e.g., $L, N, t$; vectors by lowercase boldface letters, e.g., $\mathbf{x}, \mathbf{z}$; and matrices by uppercase boldface letters, e.g., $\mathbf{X}$, $\mathbf{Z}$ and $\mathbf{W}$. Subscripts indicate matrix or vector indices, while a superscript $l \in \{1, \dots, L\}$ refers to the $l$-th autoencoder.

\subsection{Model Architecture}
\label{sec:architecture}

We use an ensemble of $L$ autoencoders to obtain multiple representations of the same input window. Each ensemble member consists of an encoder, which maps the input to a compressed feature representation, and a decoder, which aims to reconstruct the original signal. For the $l$-th member, these operations are defined as
\begin{align}
    \text{Encoder:} \qquad \mathbf{Z}^{(l)} &= f^{(l)}(\mathbf{X}), \\
    \text{Decoder:} \qquad \hat{\mathbf{X}}^{(l)} &= g^{(l)}(\mathbf{Z}^{(l)}).
\end{align}
Here, $\mathbf{X} \in \mathbb{R}^{N \times C}$ denotes the input waveform window, with $N$ timesteps and $C$ channels. The functions $f^{(l)}$ and $g^{(l)}$ denote the encoder and decoder, respectively. All ensemble members share the same architecture but have independently initialized parameters. The matrix $\mathbf{Z}^{(l)} \in \mathbb{R}^{N_z \times C_z}$ is the latent representation, and $\hat{\mathbf{X}}^{(l)} \in \mathbb{R}^{N \times C}$ is the reconstructed waveform for the $l$-th autoencoder.

\begin{figure}[!t]
\centering
\resizebox{\textwidth}{!}{
\begin{tikzpicture}[
    >=Latex,
    font=\small,
    arrow/.style={->, thick},
    thinarrow/.style={->, thin},
    encoder/.style={draw=blue!60!black, fill=blue!12, rounded corners=2pt},
    bottleneck/.style={draw=green!50!black, fill=green!15, rounded corners=2pt},
    decoder/.style={draw=purple!60!black, fill=purple!12, rounded corners=2pt},
    input/.style={draw=gray!70!black, fill=gray!15, rounded corners=2pt},
    output/.style={draw=gray!70!black, fill=gray!15, rounded corners=2pt},
    residual/.style={draw=orange!70!black, fill=orange!18, rounded corners=2pt},
    inset/.style={draw=gray!70!black, fill=white, rounded corners=2pt, line width=0.4pt},
    insetconnector/.style={densely dotted, thin, gray!70!black, shorten <=1pt, shorten >=1pt},
    note/.style={align=center, font=\scriptsize}
]

\def\dx{1.20}

\newcommand{\Shape}[2]{\(\scriptstyle #1{\!\times\!}#2\)}

\newcommand{\FeatureMap}[7]{%
    \coordinate (#1c) at (#2,#3);
    \node[#6, minimum width=#4, minimum height=#5] (#1) at (#1c) {};
    \node[align=center, font=\scriptsize] at (#1) {#7};
}

\newcommand{\FeatureMapAt}[6]{%
    \coordinate (#1c) at #2;
    \node[#5, minimum width=#3, minimum height=#4] (#1) at (#1c) {};
    \node[align=center, font=\scriptsize] at (#1) {#6};
}

\FeatureMap{X}{0}{0}{0.75cm}{2.8cm}{input}{
    $\mathbf{X}$\\[-1pt]
    \Shape{3000}{3}
}

\FeatureMap{D1}{\dx}{0}{0.75cm}{2.55cm}{encoder}{
    $D_1$\\[-1pt]
    \Shape{1500}{8}
}
\FeatureMap{D2}{{2*\dx}}{0}{0.76cm}{2.25cm}{encoder}{
    $D_2$\\[-1pt]
    \Shape{750}{16}
}
\FeatureMap{D3}{{3*\dx}}{0}{0.82cm}{1.95cm}{encoder}{
    $D_3$\\[-1pt]
    \Shape{375}{32}
}
\FeatureMap{D4}{{4*\dx}}{0}{0.76cm}{1.65cm}{encoder}{
    $D_4$\\[-1pt]
    \Shape{188}{64}
}
\FeatureMap{D5}{{5*\dx}}{0}{0.78cm}{1.35cm}{encoder}{
    $D_5$\\[-1pt]
    \Shape{94}{64}
}

\draw[arrow] (X.east) -- (D1.west);
\draw[arrow] (D1.east) -- (D2.west);
\draw[arrow] (D2.east) -- (D3.west);
\draw[arrow] (D3.east) -- (D4.west);
\draw[arrow] (D4.east) -- (D5.west);

\node[note, below=0.0mm of D1] {$k=15$};
\node[note, below=0.0mm of D2] {$k=13$};
\node[note, below=0.0mm of D3] {$k=11$};
\node[note, below=0.0mm of D4] {$k=9$};
\node[note, below=0.0mm of D5] {$k=7$};

\FeatureMap{R}{{6*\dx}}{0}{0.78cm}{1.35cm}{residual}{
    $5{\times}$ Res\\[-1pt]
    \Shape{94}{64}
}

\draw[arrow] (D5.east) -- (R.west);
\node[note, below=0.0mm of R] {$k=5$};

\FeatureMap{U1}{{7*\dx}}{0}{0.78cm}{1.65cm}{decoder}{
    $U_1$\\[-1pt]
    \Shape{188}{32}
}
\FeatureMap{U2}{{8*\dx}}{0}{0.76cm}{1.95cm}{decoder}{
    $U_2$\\[-1pt]
    \Shape{375}{16}
}
\FeatureMap{U3}{{9*\dx}}{0}{0.76cm}{2.25cm}{decoder}{
    $U_3$\\[-1pt]
    \Shape{750}{8}
}
\FeatureMap{U4}{{10*\dx}}{0}{0.70cm}{2.55cm}{decoder}{
    $U_4$\\[-1pt]
    \Shape{1500}{4}
}
\FeatureMap{U5}{{11*\dx}}{0}{0.75cm}{2.8cm}{decoder}{
    $U_5$\\[-1pt]
    \Shape{3000}{3}
}

\FeatureMap{Xhat}{{12*\dx}}{0}{0.75cm}{2.8cm}{output}{
    $\hat{\mathbf{X}}^{(l)}$\\[-1pt]
    \Shape{3000}{3}
}

\draw[arrow] (R.east) -- (U1.west);
\draw[arrow] (U1.east) -- (U2.west);
\draw[arrow] (U2.east) -- (U3.west);
\draw[arrow] (U3.east) -- (U4.west);
\draw[arrow] (U4.east) -- (U5.west);
\draw[arrow] (U5.east) -- (Xhat.west);

\node[note, below=0.0mm of U1] {$k=7$};
\node[note, below=0.0mm of U2] {$k=9$};
\node[note, below=0.0mm of U3] {$k=11$};
\node[note, below=0.0mm of U4] {$k=13$};
\node[note, below=0.0mm of U5] {$k=15$};

\coordinate (latenttap) at ($($(R.east)!0.38!(U1.west)$)+(-0.10,0)$);

\FeatureMapAt{Z}{($(latenttap)+(0,-2.55)$)}{0.78cm}{1.05cm}{bottleneck}{
    $\mathbf{Z}^{(l)}$\\[-1pt]
    \Shape{94}{64}
}

\draw[thinarrow] (latenttap) -- (Z.north);


\coordinate (InsetRow) at ($(X.south)+(0,-1.45cm)$);

\node[inset, inner sep=1.5pt] (Xinset) at (InsetRow)
    {\includegraphics[width=2.175cm,height=1.17cm]{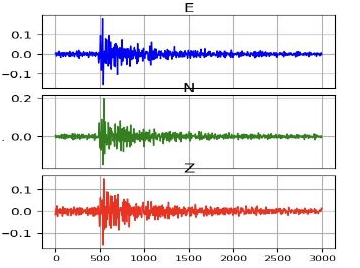}};
\draw[insetconnector] (X.south) -- (Xinset.north);
\node[note, below=1mm of Xinset] {input\\waveform};

\node[inset, inner sep=1.5pt] (Zinset) at ($(InsetRow -| Z.west)+(-1.45cm,0)$)
    {\includegraphics[width=1.875cm,height=1.17cm]{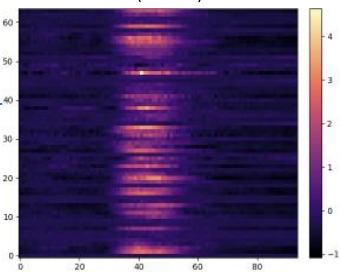}};
\draw[insetconnector] 
    (Z.west) -- ($(Z.west -| Zinset.east)$) -- (Zinset.east);
\node[note, below=1mm of Zinset] {latent\\representation};

\node[inset, inner sep=1.5pt] (Xhatinset) at (InsetRow -| Xhat)
    {\includegraphics[width=2.175cm,height=1.17cm]{figures/waveform.pdf}};
\draw[insetconnector] (Xhat.south) -- (Xhatinset.north);
\node[note, below=1mm of Xhatinset] {reconstructed\\waveform};

\coordinate (GroupTop)    at ($(X.north)+(0,0.32)$);
\coordinate (GroupBottom) at ($(X.south)+(0,-0.35)$);

\coordinate (EncNW) at ($(D1.west |- GroupTop)+(-0.12,0)$);
\coordinate (EncSE) at ($(R.east |- GroupBottom)+(0.20,0)$);

\coordinate (DecNW) at ($(U1.west |- GroupTop)+(-0.04,0)$);
\coordinate (DecSE) at ($(U5.east |- GroupBottom)+(0.20,0)$);

\node[
    draw,
    dotted,
    thick,
    rounded corners=4pt,
    fit=(EncNW)(EncSE),
    inner sep=0pt,
    label={[font=\small\bfseries, yshift=1mm]above:Encoder $f^{(l)}$}
] (EncoderBox) {};

\node[
    draw,
    dotted,
    thick,
    rounded corners=4pt,
    fit=(DecNW)(DecSE),
    inner sep=0pt,
    label={[font=\small\bfseries, yshift=1mm]above:Decoder $g^{(l)}$}
] (DecoderBox) {};

\end{tikzpicture}
}
\caption{
Overview of one member of the convolutional autoencoder ensemble.
Each member \(l\in\{1,\dots,L\}\) has the same architecture but independent
parameters, with encoder \(f^{(l)}\), decoder \(g^{(l)}\), latent representation
\(\mathbf{Z}^{(l)}\in\mathbb{R}^{94\times64}\), and reconstruction
\(\hat{\mathbf{X}}^{(l)}\). Shapes are shown as temporal length \(\times\)
channel count and $k$ represents convolutional kernel size. The blocks \(\{D_i\}_{i=1}^{5}\) denote the strided
convolutional downsampling blocks of the encoder, while
\(\{U_i\}_{i=1}^{5}\) denote the decoder upsampling/output blocks.
Representative visual insets illustrate the input waveform, latent
representation, and reconstructed waveform.
}
\label{fig:method_overview}
\alttext{A left-to-right block diagram of one autoencoder. An input block of
3000 by 3 feeds a dashed encoder group holding five downsampling blocks whose
shapes shrink from 1500 by 8 to 94 by 64 while the kernel size falls from 15 to
5, followed by a stack of five residual blocks at 94 by 64. A branch leads down
from the encoder output to the latent representation, drawn as a heat map. A
dashed decoder group of five upsampling blocks then grows the shape back from
188 by 32 to 3000 by 3, ending in a reconstructed output block. Small insets
below the diagram show the input waveform, the latent representation, and the
reconstructed waveform.}
\end{figure}

Figure~\ref{fig:method_overview} shows an overview of a single autoencoder. The network consists of downsampling, residual, and upsampling layers, detailed in Fig.~\ref{fig:cnn_autoencoder_building_blocks}. The downsampling and residual blocks form the encoder, while the upsampling blocks form the decoder. Each downsampling layer halves the temporal length of its input while potentially increasing the number of channels. Repeated downsampling compresses the temporal information, whereas the residual blocks increase the representational capacity of the bottleneck. The decoder then reconstructs the original waveform from the compressed representation. We detail each layer type below.

\paragraph{Downsampling layers} Each downsampling layer consists of reflective padding, followed by a 1-dimensional convolution with a stride of 2, batch normalization, and a ReLU activation. We apply five successive downsampling layers, which reduces the temporal length of the input waveform windows from $3000$ to $94$ timesteps. To ensure shape compatibility, we add a reflection padding step when intermediate sequence lengths are not perfectly divisible by 2.

\paragraph{Residual layers} Each residual layer comprises two identical blocks containing a 1-dimensional convolution, batch normalization, and a ReLU activation. A skip connection is added to these two blocks as indicated in Fig.~\ref{fig:cnn_autoencoder_building_blocks}. We use a stack of five such residual layers with $64$ filters and a kernel size of $5$. The final residual layer applies a linear activation instead of ReLU.

\paragraph{Upsampling layers} Each upsampling layer applies an upsampling operation followed by reflective padding, a 1-dimensional convolution, batch normalization, and a ReLU activation. Five such layers are used in succession to expand the temporal length from $94$ back to $3000$ timesteps. To maintain exact shape compatibility with the original input waveform window, we add cropping layers where necessary.

\begin{figure}[!t]
\centering
\resizebox{0.72\textwidth}{!}{
\begin{tikzpicture}[
    font=\scriptsize,
    >=Latex,
    conn/.style={
        draw,
        line width=0.55pt,
        -{Latex[length=1.4mm,width=1.0mm]}
    },
    skip/.style={
        draw,
        line width=0.55pt,
        rounded corners=7pt,
        -{Latex[length=1.4mm,width=1.0mm]}
    },
    fmap/.style={
        draw,
        rounded corners=2pt,
        minimum width=2.85cm,
        minimum height=0.50cm,
        align=center,
        fill=gray!12
    },
    iofmap/.style={
        fmap,
        draw=gray!65!black,
        fill=gray!12
    },
    layer/.style={
        draw,
        rounded corners=1.5pt,
        minimum width=2.625cm,
        minimum height=0.34cm,
        align=center,
        fill=white
    },
    conv/.style={layer, draw=blue!60!black, fill=blue!8},
    bn/.style={layer, draw=orange!70!black, fill=orange!8},
    act/.style={layer, draw=green!50!black, fill=green!10},
    up/.style={layer, draw=purple!60!black, fill=purple!8},
    sum/.style={
        draw,
        circle,
        minimum size=4.8mm,
        inner sep=0pt,
        fill=gray!10
    },
    blocktitle/.style={font=\bfseries\footnotesize, align=center}
]

\newcommand{\MiniFeature}[5]{
    \foreach \s/\op in {0.08/0.28,0.04/0.55}{
        \node[#4, opacity=\op] at ($(#2,#3)+(\s,\s)$) {};
    }
    \node[#4] (#1) at (#2,#3) {#5};
}

\node[blocktitle] at (0,1.75) {Downsampling};

\MiniFeature{d_in}{0}{1.05}{iofmap}{Input}

\node[conv] (d_conv) at (0,0.30) {Conv1D, stride $2$};
\node[bn]   (d_bn)   at (0,-0.30) {BatchNorm};
\node[act]  (d_act)  at (0,-0.90) {ReLU};

\MiniFeature{d_out}{0}{-1.70}{iofmap}{Output\\{\scriptsize (temporal length halved)}}

\draw[conn] (d_in.south) -- (d_conv.north);
\draw[conn] (d_conv.south) -- (d_bn.north);
\draw[conn] (d_bn.south) -- (d_act.north);
\draw[conn] (d_act.south) -- (d_out.north);

\node[
    draw=blue!60!black,
    rounded corners=4pt,
    dashed,
    fit=(d_in)(d_conv)(d_bn)(d_act)(d_out),
    inner sep=6pt
] {};

\node[blocktitle] at (5.8,1.75) {Residual};

\MiniFeature{r_in}{5.8}{1.05}{iofmap}{Input}

\node[conv] (r_conv1) at (5.8,0.30) {Conv1D, stride $1$};
\node[bn]   (r_bn1)   at (5.8,-0.30) {BatchNorm};
\node[act]  (r_relu1) at (5.8,-0.90) {ReLU};
\node[conv] (r_conv2) at (5.8,-1.50) {Conv1D, stride $1$};
\node[bn]   (r_bn2)   at (5.8,-2.10) {BatchNorm};
\node[sum]  (r_add)   at (5.8,-2.80) {$+$};
\node[act]  (r_act)   at (5.8,-3.45) {ReLU / linear};

\MiniFeature{r_out}{5.8}{-4.25}{iofmap}{Output\\{\scriptsize (same shape with input)}}

\draw[conn] (r_in.south) -- (r_conv1.north);
\draw[conn] (r_conv1.south) -- (r_bn1.north);
\draw[conn] (r_bn1.south) -- (r_relu1.north);
\draw[conn] (r_relu1.south) -- (r_conv2.north);
\draw[conn] (r_conv2.south) -- (r_bn2.north);
\draw[conn] (r_bn2.south) -- (r_add.north);
\draw[conn] (r_add.south) -- (r_act.north);
\draw[conn] (r_act.south) -- (r_out.north);

\coordinate (r_skip_start) at ($(r_in.south)!0.2!(r_conv1.north)$);
\coordinate (r_skip_elbow) at ($(r_skip_start)+(1.50,0)$);
\draw[skip] (r_skip_start) -- (r_skip_elbow) |- (r_add.east);

\node[
    draw=orange!70!black,
    rounded corners=4pt,
    dashed,
    fit=(r_in)(r_conv1)(r_bn1)(r_relu1)(r_conv2)(r_bn2)(r_add)(r_act)(r_out),
    inner sep=6pt
] {};

\node[blocktitle] at (11.6,1.75) {Upsampling};

\MiniFeature{u_in}{11.6}{1.05}{iofmap}{Input}

\node[up]   (u_up)   at (11.6,0.30) {Upsample};
\node[conv] (u_conv) at (11.6,-0.30) {Conv1D, stride $1$};
\node[bn]   (u_bn)   at (11.6,-0.90) {BatchNorm};
\node[act]  (u_act)  at (11.6,-1.50) {ReLU};

\MiniFeature{u_out}{11.6}{-2.30}{iofmap}{Output\\{\scriptsize (temporal length doubled)}}

\draw[conn] (u_in.south) -- (u_up.north);
\draw[conn] (u_up.south) -- (u_conv.north);
\draw[conn] (u_conv.south) -- (u_bn.north);
\draw[conn] (u_bn.south) -- (u_act.north);
\draw[conn] (u_act.south) -- (u_out.north);

\node[
    draw=purple!60!black,
    rounded corners=4pt,
    dashed,
    fit=(u_in)(u_up)(u_conv)(u_bn)(u_act)(u_out),
    inner sep=6pt
] {};

\end{tikzpicture}
}
\caption{
Visualization of the convolutional building blocks used in the autoencoder. Autoencoders are composed of a compressive encoder that is composed of downsampling and residual blocks, and upsampling blocks for decoding the latent representation. Downsampling blocks halve the temporal resolution through strided convolution, residual blocks increase depth while preserving feature-map shape, and upsampling blocks doubles the temporal resolution.
}
\label{fig:cnn_autoencoder_building_blocks}
\alttext{Three vertical flow charts side by side, each a dashed box of labeled
boxes joined by downward arrows. The downsampling chart runs input, one
dimensional convolution with stride 2, batch normalization, ReLU, and an output
of halved temporal length. The residual chart runs input, convolution with
stride 1, batch normalization, ReLU, a second convolution and batch
normalization, then a circled addition node that also receives a skip
connection drawn from the input, followed by a ReLU or linear activation and an
output of the same shape as the input. The upsampling chart runs input,
upsample, convolution with stride 1, batch normalization, ReLU, and an output
of doubled temporal length.}
\end{figure}

\subsection{Ensemble Training Objective}
\label{sec:training_objective}

The training objective consists of two main components: the reconstruction loss and the alignment loss. The reconstruction loss is standard for autoencoders~\cite{goodfellow2016deep} and ensures that the learned latent representations retain sufficient information to reconstruct the original waveform.

The alignment loss is specific to our method and encourages the linear maps to emphasize shared temporal modes of variation across independently learned latent representations. For noise waveform windows, the latent representations generally exhibit weak temporal alignment, and they are therefore attenuated by the learned linear maps. In contrast, event waveform windows produce latent representations with strong shared modes of temporal variation. These modes are further emphasized by the linear maps, leading to prominent cross-covariances between the transformed representations.

\subsubsection{Reconstruction Loss}
The reconstruction loss is defined as the ensemble-averaged root mean squared error (RMSE) between the input waveform window and reconstructed waveform windows of each autoencoder. Thus,
\begin{equation}
    \mathcal{L}_{\mathrm{reconstruction}}
    =
    \frac{1}{L}
    \sum_{l=1}^{L}
    \operatorname{RMSE}
    \left(
        \mathbf{X},
        \hat{\mathbf{X}}^{(l)}
    \right),
\end{equation}
where $L$ is the number of autoencoders in the ensemble, and $\mathbf{x}$ and $\hat{\mathbf{x}}^{(l)}$ denote the temporally
demeaned input and reconstruction, respectively. The RMSE is computed as
\begin{equation}
    \operatorname{RMSE}
    \left(
        \mathbf{X},
        \hat{\mathbf{X}}^{(l)}
    \right)
    =
    \sqrt{
        \frac{1}{NC}
        \sum_{t=1}^{N}
        \sum_{c=1}^{C}
        \left(
            x_{t,c}
            -
            \hat{x}^{(l)}_{t,c}
        \right)^2
    },
\end{equation}
where $N = 3000$ is the waveform window temporal length and $C = 3$ is the number of channels.

\subsubsection{Alignment Loss}

To compare the representations learned by different ensemble members, we introduce linear alignment maps $\mathbf{W}^{(l)} \in \mathbb{R}^{C_z \times C_z}$. For each member,
\begin{equation}
    \hat{\mathbf{z}}^{(l)}_t = \mathbf{W}^{(l)} \mathbf{z}^{(l)}_t,
\end{equation}
where $\mathbf{z}^{(l)}_{t}, \hat{\mathbf{z}}^{(l)}_t \in \mathbb{R}^{C_z}$ are the original and aligned latent vectors at timestep $t$. To train these maps, we first normalize the aligned features along the temporal dimension. For a given aligned feature matrix $\hat{\mathbf{Z}}^{(l)}$, the normalized matrix $\tilde{\mathbf{Z}}^{(l)}$ is obtained by centering and $L_2$-normalizing each channel independently:
\begin{equation}
    \tilde{z}^{(l)}_{t, c} = \frac{\hat{z}^{(l)}_{t, c} - \mu^{(l)}_{c}}{\sqrt{\sum_{\tau=1}^{N_z} (\hat{z}^{(l)}_{\tau, c} - \mu^{(l)}_c)^2}}
\end{equation}
where $\mu^{(l)}_{c}$ is the temporal mean for channel $c$ of the $l$-th latent representation. We then calculate the Root Mean Square Error (RMSE) across the temporal and channel dimensions as a distance metric between two normalized representations:
\begin{equation}
    \mathcal{D}(\hat{\mathbf{Z}}^{(l)}, \hat{\mathbf{Z}}^{(k)}) = \sqrt{ \frac{1}{N_z C_z} \sum_{t=1}^{N_z} \sum_{c=1}^{C_z} \left( \tilde{z}^{(l)}_{t, c} - \tilde{z}^{(k)}_{t, c} \right)^2 }
\end{equation}
The total alignment loss minimizes the average of these pairwise distances across all $\binom{L}{2}$ unique pairs in the ensemble:
\begin{equation}
    \mathcal{L}_{\mathrm{alignment}} = \frac{2}{L(L-1)} \sum_{l=1}^{L-1} \sum_{k=l+1}^{L} \mathcal{D}(\hat{\mathbf{Z}}^{(l)}, \hat{\mathbf{Z}}^{(k)})
\end{equation}

\subsection{Detection Metric}
\label{sec:detection_metric}
The model will assign labels to each window by applying a threshold to 
a score which measures the likelihood that the window contains an event. 
This score, which we call the detection metric, is computed from the aligned latent representations $\{\hat{\mathbf{Z}}^{(l)}\}_{l=1}^{L}$. A baseline detection score can be obtained by averaging the zero-lag cross-covariances between these representations across ensemble pairs. We additionally account for the temporal uncertainty introduced by the large receptive fields of the convolutional encoders. Because the autoencoders are initialized independently, similar features may appear at slightly different latent timesteps across ensemble members. We therefore compute a weighted average of cross-covariances over small temporal shifts, which produces the final detection metric:
\begin{equation}
    m(\mathbf{X}) = \frac{1}{L(L-1)} \sum_{s=-S}^{S} \sum_{l=1}^{L} \sum_{\substack{k=1 \\ k \neq l}}^{L} w_s \, \mathrm{cov}\!\left(\operatorname{shift}_s(\hat{\mathbf{Z}}^{(l)}), \hat{\mathbf{Z}}^{(k)}\right)
\end{equation}
where $\operatorname{shift}_s(\hat{\mathbf{Z}}^{(l)})$ denotes a circular temporal shift of the aligned latent feature matrix by $s$ latent timesteps, and $\mathrm{cov}$ denotes the channel-averaged cross-covariance. The shift weights are defined as
\begin{equation}
    w_s
    =
    \frac{
        \exp\left(-s^2 / 2\sigma_S^2\right)
    }{
        \sum_{s'=-S}^{S}
        \exp\left(-{s'}^2 / 2\sigma_S^2\right)
	} .
\end{equation}
We set $\sigma_S = 8$ latent timesteps. Since the encoder maps a 30\,s waveform to approximately 94 latent timesteps, this weighting scale corresponds to about $2.5$\,s in the input waveform.

\subsection{Training Procedure}
\label{sec:training_procedure}

\subsubsection{Optimization Details}

For all experiments, model parameters were optimized using Adam with a constant learning rate of $10^{-3}$, $\epsilon=10^{-7}$, and momentum parameters $\beta_1=0.99$ and $\beta_2=0.999$. The batch size was set to 256. All other optimizer settings were kept at their default values; in particular, we did not use exponential moving average filtering, weight decay, or gradient clipping.

During training, we record model snapshots at each epoch and select the snapshot with the lowest validation loss. Unlike in supervised training, this validation loss only measures how well the unsupervised training objective is satisfied. To reflect a realistic deployment setting in which class labels are unavailable, no classification labels or detection performance metrics are used in the model selection procedure.

\subsubsection{Standard Mini-Batch Training}

For datasets in which event and noise waveform windows are sufficiently represented, the model is trained using standard mini-batch optimization. At each training step, a mini-batch of $B$ waveform windows is sampled from the training set, and all waveform windows in the mini-batch contribute to the reconstruction and alignment losses defined above. In contrast to the score-based batch dilation procedure described below, standard mini-batch training does not perform any intermediate scoring or sample selection before the gradient update.

\subsubsection{Batch Dilation for Imbalanced Continuous Data}
\label{sec:dilation}

When training on continuous seismic recordings, severe class imbalance can arise because event-containing windows typically constitute only a small fraction of the data. Under these conditions, the alignment loss can be dominated by noise, causing the learned linear maps to align noise-related features rather than event-related structure, which can degrade detection performance.

We address this problem using batch dilation. At each training step, the current model is first applied in inference mode to assign a detection score to each candidate waveform window. Instead of updating the model using a single mini-batch directly, we temporarily merge multiple mini-batches to form a larger candidate waveform pool, and drop the low-scoring waveform windows in this larger set to form a single batch of high-scoring windows. Only this batch is used for computing gradients. Consequently, the batches used for training contain a higher concentration of event waveforms in comparison to their natural occurrence, increasing their influence over on the optimization dynamics.

The number of mini-batches merged at each training step is controlled by the dilation number $D(e)$, which increases during training. Let $B$ denote the number of waveform windows used for each gradient update, let $E$ denote the total number of training epochs, and let $e \in \{1,\ldots,E\}$ be the epoch index. We set
\begin{equation}
    D(e)
    =
    \operatorname{round}
    \left(
        D_{\max}^{\,e/E}
    \right),
    \label{eq:dilation_log}
\end{equation}
where $D_{\max}$ is the final dilation number. At epoch $e$, we merge $D(e)$ mini-batches, score all waveform windows in the merged set, and retain only the top $B$ waveform windows for the gradient update. Thus, $D(1)=1$ corresponds to standard mini-batch training, while larger values of $D(e)$ make the update increasingly focused on windows that score as event-like.

Batch dilation is related to sample-selection methods such as sample propagation~\cite{ml:garcia2026rareaware} and online hard-example mining~\cite{ml:shrivastava2016ohem}. These are supervised methods that
train on the highest-loss samples in order to focus learning where the model performs worst, whereas batch dilation is an unsupervised method that
focus learning on the examples that are more likely to contain an event according to the method's own measure, without 
having access to any labels. As training proceeds, the updated model re-scores the windows using its latest weights. Thus, it can be called a form of rare-example mining.

\section{Experiments}
\label{sec:experiments}
Throughout this section, we use the term \emph{seismic event} to denote the positive class, i.e., cataloged earthquakes in the benchmark and regional datasets. A waveform window containing a relevant phase arrival is referred to as an \emph{event window}, while all other windows are referred to as \emph{noise windows}.

To test generalization, we evaluate the model in two settings. In \emph{in-domain evaluation}, the training and test sets are non-overlapping subsets of the same dataset. In \emph{cross-domain evaluation}, the model is trained on one dataset and evaluated on another. 

To compare our method with supervised counterparts, we first run experiments on public datasets with high event-window fractions, namely STEAD (88.6\% event windows)~\cite{mousavi2019stead} and INSTANCE (81.4\% event windows)~\cite{michelini2021instance}. We then focus on continuous seismic recordings, in which most waveform windows are dominated by noise and event-window fractions are below 1\%. For this purpose, we prepare two datasets covering two recent seismic activities in the Marmara region of Türkiye: the 2019 Silivri sequence~\cite{schmittbuhl2016seismicity, Karabulut2020MarmaraCatalog} and the 2025 Sindirgi sequence, which we refer to as the Silivri and Bal{\i}kesir--S{\i}nd{\i}rg{\i} datasets, respectively (for preparation details, see Section~\ref{sec:raw-datasets-and-training}). This setting tests whether the model can be trained on highly imbalanced datasets and used to learn region-specific detectors directly from available waveform archives, without requiring labels.

Finally, we evaluate the method in a complementary role by using it to improve the precision of PhaseNet. This tests whether the unsupervised score provides information that is not fully captured by supervised phase probabilities.

\subsection{Overview}\label{sec:overview}
\subsubsection{Preprocessing}\label{sec:preprocessing}
Our experiments involve waveforms recorded by different instruments at various sampling rates, whereas RECOVAR requires a consistent sampling rate of 100 Hz. The public benchmark datasets, STEAD and INSTANCE, contain waveforms already sampled at this rate and therefore require no additional resampling. For the continuous-waveform datasets, namely Silivri and Bal{\i}kesir--S{\i}nd{\i}rg{\i}, recordings are resampled to 100 Hz when necessary.

For all datasets, we use the east--west (\texttt{E}), north--south (\texttt{N}), and vertical (\texttt{Z}) components as model inputs and arrange them in a consistent \texttt{E--N--Z} order. We do not restrict the sensor type or explicitly distinguish between accelerometer and seismometer recordings.

To obtain fixed-length 30 s model inputs, we first crop each waveform to an intermediate preprocessing window of 40 s, unless otherwise stated. Each channel is then linearly detrended and bandpass filtered between 1 and 20 Hz. We retain the central 30 s segment as the model input, thereby reducing edge effects introduced by filtering. Finally, each channel is normalized independently.

\subsubsection{Optimization and Training Settings}\label{sec:optimization}
In all experiments, we use the optimization and checkpoint-selection procedure described in Section~\ref{sec:training_procedure}.

In experiments without batch dilation, we train the model for 20 epochs. For the continuous waveforms where we use batch dilation, we train for 100 epochs for Silivri (where the maximum dilation rate is $D_{\max} = 64$), whereas for Bal{\i}kesir--S{\i}nd{\i}rg{\i} we train for 40 epochs using $D_{\max} = 8$.\footnote{As detailed in Section~\ref{sec:dilation}, $D_{\max}$ corresponds to the number of batches used to obtain a single training batch in the final epoch.} 

Training is conducted on a workstation equipped with a single NVIDIA RTX 3090 Ti GPU. Training takes approximately 6.5 and 7.5 minutes per epoch for the INSTANCE and STEAD datasets, respectively.

\subsubsection{Evaluation Metrics}\label{sec:evaluation-metrics}
\paragraph{Detection outcomes}
We evaluate detections by matching them to cataloged seismic events. A detection matched to a cataloged seismic event is counted as a true positive ($TP$), whereas an unmatched detection is counted as a false positive ($FP$). A cataloged seismic event with no matched detection is counted as a false negative ($FN$), and a waveform window correctly identified as noise is counted as a true negative ($TN$). These definitions follow the standard confusion-matrix terminology for binary classification~\cite{fawcett2006roc,powers2011evaluation}.

From these quantities, the true positive rate (TPR, or recall) and false positive rate (FPR) are defined as
\[
\mathrm{TPR} = \frac{TP}{TP+FN}, \qquad
\mathrm{FPR} = \frac{FP}{FP+TN}.
\]

\paragraph{ROC-AUC score}
Detection algorithms typically rely on an adjustable threshold to determine whether a waveform window is classified as an event window or a noise window. As this threshold varies, the true positive rate and false positive rate change accordingly. The Receiver Operating Characteristic (ROC) curve summarizes this behavior by plotting TPR against FPR across all threshold values. The Area Under the ROC Curve (ROC-AUC) provides a threshold-independent summary of the model's ability to distinguish between the event and noise classes~\cite{fawcett2006roc}. Unless stated otherwise, we therefore use ROC-AUC as the primary metric for waveform-window classification, consistent with previous evaluations of seismic detection models~\cite{rev:munchmeyer2022picker}. A perfect classifier has a ROC-AUC score of 1.0.

\paragraph{Precision and recall}
When using RECOVAR in conjunction with PhaseNet, we assess the performance improvement using precision and recall, defined as follows:
\[
\mathrm{Precision} = \frac{TP}{TP + FP}, \qquad
\mathrm{Recall} = \frac{TP}{TP + FN} .
\]

Precision is the proportion of true seismic events among all positive label assignments made by the system, while recall is the fraction of ground-truth catalog events that are detected~\cite{powers2011evaluation}. Varying the decision threshold generally produces a trade-off between these quantities and traces a precision--recall curve, which is particularly informative for imbalanced detection problems~\cite{davis2006precision}. 

\paragraph{$F_1$ Score}
To summarize precision and recall using a single metric, we use the $F_1$ score, defined as their harmonic mean~\cite{powers2011evaluation}:
\[
F_1 = 2 \, \frac{\mathrm{Precision} \times \mathrm{recall}}{\mathrm{Precision} + \mathrm{Recall}}.
\]
In Section~\ref{sec:postfilter}, we select the threshold that maximizes $F_1$ to identify the operating point providing the best balance between retaining cataloged seismic events and suppressing false PhaseNet detections.

\subsection{Training and Classification on Public Datasets}
\label{sec:benchmark_results}

We first evaluate the method in a standard benchmark setting by comparing it with supervised seismic detection models on labeled waveform datasets. Although the method is trained without event labels, these labeled benchmarks allow us to measure how well its detection scores separate event and noise waveforms. We consider both in-domain and cross-domain evaluation.

To ensure the closest possible comparison, we adopt the benchmark protocol of M\"unchmeyer \emph{et al.} \cite{rev:munchmeyer2022picker}, which evaluates supervised models under both in-domain and cross-domain settings. We follow their preprocessing, train--test split construction, and ROC-AUC evaluation procedure, while replacing the supervised learning objective with the unsupervised training procedure described above.

\subsubsection{Datasets and Cross-Validation}
\label{sec:datasets-and-cross-validation}

We use two public datasets, STEAD~\cite{mousavi2019stead} and INSTANCE~\cite{michelini2021instance}, to evaluate in-domain and cross-domain performance. Because these datasets originate from different geographic regions and use different limits on epicentral distance, they allow us evaluate performance under distributional shifts.

Both datasets use three channels—East (E), North (N), and Vertical (Z)—and share a sampling rate of 100 Hz. INSTANCE contains 1,159,249 event waveforms and 132,330 noise waveforms, whereas STEAD contains 1,050,000 event waveforms and 100,000 noise waveforms. INSTANCE provides 120 s waveform windows and covers epicentral distances up to 600 km, while STEAD provides 60 s windows and is restricted to distances below 350 km. Geographically, STEAD has global coverage, whereas INSTANCE focuses on Italy.

\paragraph{Training set preparation}
Following the benchmark protocol of~\cite{rev:munchmeyer2022picker}, we randomly crop two thirds of the event waveforms such that at least one phase arrival is guaranteed to fall within the window. The remaining event waveforms and all noise waveforms are cropped randomly without this constraint.

We employ 5-fold cross-validation on both STEAD and INSTANCE to obtain a robust estimate of performance. For each fold, we select the checkpoint with the lowest unsupervised validation loss, defined as the sum of the reconstruction and alignment losses. No event labels or detection metrics are used for model selection.

\paragraph{Test set preparation}
For testing, event waveforms are cropped such that the onset time falls within the window with a 3 s margin. Test examples that do not allow this margin are discarded. We use 60\% of the dataset for training, consistent with M\"unchmeyer \emph{et al.} \cite{rev:munchmeyer2022picker}. However, to perform five-fold cross-validation and quantify performance variability across folds, we allocate the remaining data equally between validation and testing, using 20\% for each, rather than the 10\% validation and 30\% test split used by M\"unchmeyer \emph{et al.} \cite{rev:munchmeyer2022picker}.

\subsubsection{Results}\label{results}  

\begin{table}[!t]
\renewcommand{\arraystretch}{1.15}
\centering
\caption{ROC-AUC scores for different training datasets (rows) and testing datasets (columns). A perfect classifier has an ROC-AUC of 1.0, whereas a random classifier has an ROC-AUC of 0.5. RECOVAR results are reported as the mean $\pm$ standard deviation across the $K=5$ cross-validation folds, with the model trained and evaluated using the corresponding training and test split in each fold (for details, see
Section~\ref{sec:datasets-and-cross-validation}). PhaseNet and EQTransformer scores are taken from M\"unchmeyer \emph{et al.} \cite{rev:munchmeyer2022picker}. All scores therefore correspond to the training and evaluation protocol of that study, and the comparison should be read within that protocol. The best score in each training--testing combination is shown in bold.}
\label{tab:results}
\begin{tabular}{|l|l|c|c|}
\hline
\multirow{2}{*}{\makecell{\textbf{Training}\\ \textbf{dataset}}}
& \multirow{2}{*}{\textbf{Model}}
& \multicolumn{2}{c|}{\makecell{\textbf{Testing}\\ \textbf{dataset}}} \\
\cline{3-4}
& & \textbf{INSTANCE} & \textbf{STEAD} \\
\hline
\multirow{3}{*}{INSTANCE}
& Ours               & $\mathbf{0.976 \pm 0.001}$ & $0.988 \pm 0.001$   \\
& PhaseNet              & $0.964$                    & $\mathbf{0.994}$    \\
& EQTransformer         & $0.957$                    & $0.990$             \\
\hline
\multirow{3}{*}{STEAD}
& Ours               & $\mathbf{0.974 \pm 0.001}$ & $0.988 \pm 0.001$   \\
& PhaseNet              & $0.941$                    & $\mathbf{1.000}$    \\
& EQTransformer         & $0.966$                    & $\mathbf{1.000}$    \\
\hline
\end{tabular}

\end{table}

We compare the method with PhaseNet and EQTransformer in both in-domain and cross-domain settings, following the benchmark protocol of M\"unchmeyer \emph{et al.} \cite{rev:munchmeyer2022picker}. We observe that the method maintains competitive performance while remaining more stable across datasets. Its ROC-AUC ranges from $0.974 \pm 0.001$ to $0.988 \pm 0.001$ across all training--testing combinations, whereas PhaseNet and EQTransformer show greater variation, with scores ranging from $0.941$ to $1.000$ and from $0.957$ to $1.000$, respectively.

The results also suggest that the method has performance characteristics complementary to those of supervised methods. RECOVAR achieves a higher ROC-AUC on INSTANCE, which includes events at larger epicentral distances, whereas the supervised models perform consistently better on STEAD, where events generally have clearer onsets.

\subsection{Training and Classification on Continuous Seismic Waveforms}\label{sec:benchmark_results_continuous}

Unsupervised learning removes labeling bias and the need to curate a labeled dataset. This allows us to train our model on continuous seismic waveforms. 

After applying the standard preprocessing described in Section~\ref{sec:preprocessing}, we directly use continuous seismic waveforms obtained from stations in two regions in Türkiye, described below. These datasets differ from the benchmark datasets (STEAD and INSTANCE): the fraction of waveform windows labeled as events is below 1\%, making this a more challenging setting.

\subsubsection{Dataset Preparation}\label{sec:raw-datasets-and-training}
To handle continuous data, we partition each station's record into 60 s waveform windows. After applying the preprocessing pipeline described in Section~\ref{sec:preprocessing}, windows with data gaps are discarded and the central 30 s waveform window is retained. This construction helps avoid having samples from the same event in successive windows, although some noise-labeled windows may still contain residual coda from preceding seismic events. To obtain ground-truth labels, we use cataloged P-wave arrival times for each station. A waveform window is labeled as an \emph{event window} if a P-wave arrival falls within its 30 s interval; otherwise, it is labeled as a \emph{noise window}.

\paragraph{Silivri}\label{sec:silivri}
We use continuous three-component records from 17 KOERI broadband stations distributed around the Sea of Marmara (BGKT, CTKS, ERIK, GELI, GONE, ISK, IZI, KCTX, KLYT, MDNY, MRMT, ORLT, OSMT, SLVT, TKR, UKOP and YLV), spanning 1 September to 30 November 2019. The stations span approximately $26.4^\circ$--$29.7^\circ$E and $40.0^\circ$--$41.3^\circ$N. The catalog lists 523 seismic events in the study area over this period~\cite{durand2020twoscale}. These events form a single compact offshore cluster near $28.2^\circ$E, $40.9^\circ$N, corresponding to the 2019 Silivri swarm, so the sources are concentrated in a small area relative to the aperture of the recording network. After window construction, the dataset contains 3{,}291 event windows out of 2{,}075{,}648 total windows, corresponding to a $0.16\%$ positive-window rate.

\paragraph{Bal{\i}kesir--S{\i}nd{\i}rg{\i}}\label{sec:balikesir}
We use continuous three-component records from six KOERI broadband stations in the Bal{\i}kesir--S{\i}nd{\i}rg{\i} region (BALB, GORD, KULA, SIMA, SOMA and SUSR), spanning approximately $27.5^\circ$--$29.0^\circ$E and $38.5^\circ$--$40.0^\circ$N. The catalog spans from 1 July 2025 to 28 February 2026, covering the 2025 Bal{\i}kesir--S{\i}nd{\i}rg{\i} sequence~\cite{sozbilir2026seismic}. The KOERI catalog lists 3{,}048 seismic events in the study area over this period. Most of these events fall in an elongated, roughly east--west cluster near $28.0^\circ$--$28.5^\circ$E, $39.1^\circ$--$39.3^\circ$N, enclosed by the central stations, with a smaller secondary cluster about 30 km to the east. Compared with Silivri, the sources are therefore both more numerous and more spatially distributed across the network. After window construction, the dataset contains 15{,}559 event windows out of 2{,}039{,}040 total windows, corresponding to a $0.76\%$ positive-window rate.

\subsubsection{Results}\label{sec:continuous-results}

We ran the continuous waveform experiments both with and without batch dilation. Table~\ref{tab:dilation_ablation} reports the ROC-AUC scores obtained under in-domain and cross-domain evaluation. The effect of batch dilation is highest when RECOVAR is trained on the Silivri dataset, which has the lowest positive-window rate among the training datasets. In this case, the improvement is consistent across all testing datasets, reaching $+0.073$ on INSTANCE, $+0.252$ on Silivri, and $+0.034$ on Bal{\i}kesir--S{\i}nd{\i}rg{\i}. The largest gain is observed in the in-domain Silivri evaluation, where the ROC-AUC score increases from $0.698$ to $0.950$.

\begin{table}[!ht]
\renewcommand{\arraystretch}{1.15}
\scriptsize
\setlength{\tabcolsep}{2.5pt}
\centering

\caption{RECOVAR ROC-AUC scores under in-domain and cross-domain evaluation.
Rows indicate the testing dataset, while columns indicate the training dataset
and training configuration. For the Silivri and
Bal{\i}kesir--S{\i}nd{\i}rg{\i} training datasets, $\Delta$ denotes the change
obtained by applying batch dilation.}
\label{tab:dilation_ablation}
\begin{tabular*}{\textwidth}
{@{\extracolsep{\fill}}|c|c|ccc|ccc|}
\hline
\textbf{Testing dataset}
& \multicolumn{7}{c|}{\textbf{Training dataset}} \\
\cline{2-8}
&
\textbf{INSTANCE}
& \multicolumn{3}{c|}{\textbf{Silivri}}
& \multicolumn{3}{c|}{\textbf{Bal{\i}kesir--S{\i}nd{\i}rg{\i}}} \\
\cline{2-8}
&
\textbf{Standard}
& \textbf{No dilation}
& \textbf{Dilation}
& \boldmath{$\Delta$}
& \textbf{No dilation}
& \textbf{Dilation}
& \boldmath{$\Delta$} \\
\hline
INSTANCE
& \textbf{0.977}
& 0.896 & 0.969 & +0.073
& 0.973 & 0.970 & -0.003 \\
Silivri
& \textbf{0.977}
& 0.698 & 0.950 & +0.252
& 0.965 & 0.961 & -0.004 \\
Bal{\i}kesir--S{\i}nd{\i}rg{\i}
& \textbf{0.996}
& 0.960 & 0.994 & +0.034
& 0.995 & \textbf{0.996} & +0.001 \\
\hline
\end{tabular*}%

\end{table} 

The INSTANCE-trained model provides the highest performance, achieving ROC-AUC scores of $0.977$ on both INSTANCE and Silivri, and $0.996$ on Bal{\i}kesir--S{\i}nd{\i}rg{\i}. This is expected, since the INSTANCE dataset has a much higher positive-window rate than the continuous datasets (INSTANCE: $88.6\%$ compared with Bal{\i}kesir--S{\i}nd{\i}rg{\i}: $0.76\%$ and Silivri: $0.16\%$). Nevertheless, the model trained on Bal{\i}kesir--S{\i}nd{\i}rg{\i} remains competitive in the in-domain setting, reaching $0.996$ with batch dilation and matching the INSTANCE-trained model on the same testing dataset. 

Overall, these results show that batch dilation is particularly useful when training on highly imbalanced continuous waveform data. Although event density remains an important factor for model performance, the results suggest that the performance gap associated with low positive-window rates of continuous data can be reduced, and in some in-domain settings even closed. 
  
\subsection{Using RECOVAR as a Post-Detection Filter}
\label{sec:postfilter}

Our observations indicate that supervised models and RECOVAR generally agree on the existence of an event, but their certainty levels differ across samples. This suggests that labeling biases in supervised learning can be reduced by running supervised models at low detection thresholds to obtain high-recall candidate events, and then filtering these using an unsupervised representation-based score.

To demonstrate this complementary role, we combine RECOVAR with PhaseNet. Specifically, we apply the windowing and preprocessing pipeline described in Section~\ref{sec:preprocessing} to a regional dataset, evaluate both models on the resulting waveform windows, and use the unsupervised scores to filter the candidate detections generated using PhaseNet.

\subsubsection{Dataset Preparation}

We use nine contiguous time intervals from the Bal{\i}kesir--S{\i}nd{\i}rg{\i} region between November~3 and December~1, 2025, centered on periods of elevated seismicity (totaling 43.4~hours of continuous data per station across 6 stations). The EMSC catalog contains 183 events within these windows, with magnitudes ranging from $M_L$~0.7 to 3.9 and depths from 0.3 to 31.5 km.
    
\subsubsection{Setup}
Both RECOVAR and PhaseNet process 30 s waveform windows. Therefore, we apply the preprocessing pipeline described in Section~\ref{sec:preprocessing} to obtain 30 s preprocessed windows, and we discard any windows containing data gaps. 

We feed these waveform windows to both models. RECOVAR produces a single detection score for each window, whereas PhaseNet produces a time series of P-phase arrival probabilities. For RECOVAR, we associate each score with the timestamp at the center of the corresponding waveform window. For PhaseNet, we align the probability time series from all windows on the absolute time axis and average the probabilities corresponding to the same timestamp. This procedure gives us continuous detection-score traces for both models.

\begin{figure}[!htbp]
\centering
\includegraphics[width=0.85\textwidth]{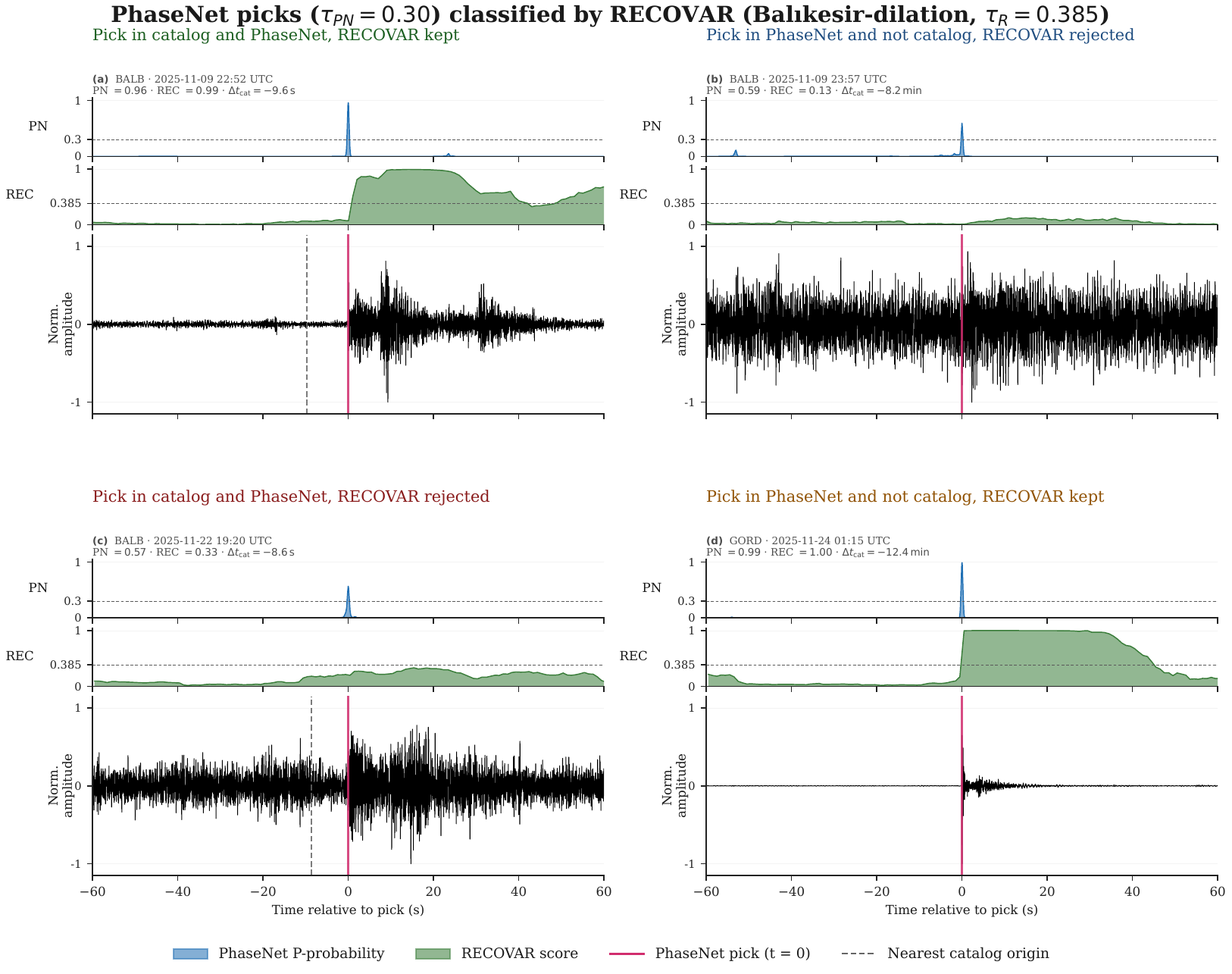}
\caption{Representative PhaseNet candidate detections obtained using a threshold of $\tau_{PN}=0.30$ and evaluated with the RECOVAR model trained on the Bal{\i}kesir--S{\i}nd{\i}rg{\i} dataset using dilation training. RECOVAR is applied with a threshold yielding the highest $F_1$ score under 3-station coincidence criterion in Table~\ref{tab:postfilter_ops}. The four panels show one example of each outcome: (a) a cataloged event detected by PhaseNet and retained by RECOVAR, (b) a PhaseNet detection with no catalog match that is rejected by RECOVAR, (c) a cataloged event detected by PhaseNet but rejected by RECOVAR, and (d) a PhaseNet detection with no catalog match that is retained by RECOVAR. Each panel shows the PhaseNet P-phase probability trace, the RECOVAR score trace, and the normalized vertical-component waveform over a 120 s window centered on the PhaseNet P-phase detection. The dashed vertical line marks the nearest cataloged origin time. The RECOVAR score trace is shifted by $+15$ s so that its rising edge indicates when the event begins to enter the corresponding waveform window, rather than the center time to which the score is originally assigned.}
\label{fig:panels_grid}
\end{figure}

PhaseNet detections are extracted from the P-probability traces using local peak finding, with a minimum separation of $1.5 s$ and a P-probability threshold of $\tau_{PN}=0.30$. In our detection-filtering pipeline, each PhaseNet detection is assigned the maximum RECOVAR score among the waveform windows containing the candidate onset. The detection is retained only if this score exceeds the RECOVAR threshold $\tau_R$.

The retained station-level detections are then grouped by temporal coincidence. Detections from different stations that occur within a 20 s coincidence window are assigned to the same group. A group is declared an event candidate when it contains detections from at least $N$ stations. We analyze the performance for $N=3$, which reflects standard network-detection practice and reduces the influence of isolated single-station noise detections, and for $N=1$, which retains all station-level detections and therefore admits substantially more false detections. An event candidate is counted as a true positive only if the median onset time of its contributing station-level detections lies within 20 s of a catalog event.

We evaluate three RECOVAR variants corresponding to different training configurations: the INSTANCE-trained model described in Section~\ref{sec:benchmark_results}, the Bal{\i}kesir--S{\i}nd{\i}rg{\i}-trained model without batch dilation, and the Bal{\i}kesir--S{\i}nd{\i}rg{\i}-trained model with batch dilation, the latter two are described in Section~\ref{sec:benchmark_results_continuous}. During this evaluation, we fix the PhaseNet threshold at $\tau_{PN}=0.30$ and sweep the RECOVAR threshold over $\tau_R \in [0.0, 1.0)$. This setup keeps the event candidates detected by PhaseNet fixed, so differences in precision and recall reflect the effect of the filtering due to RECOVAR.

\subsubsection{Results}

We evaluate the detection-filtering pipeline using precision and recall, as defined in Section~\ref{sec:evaluation-metrics}.

\begin{figure}[!htbp]
    \centering
    \includegraphics[width=0.85\textwidth]{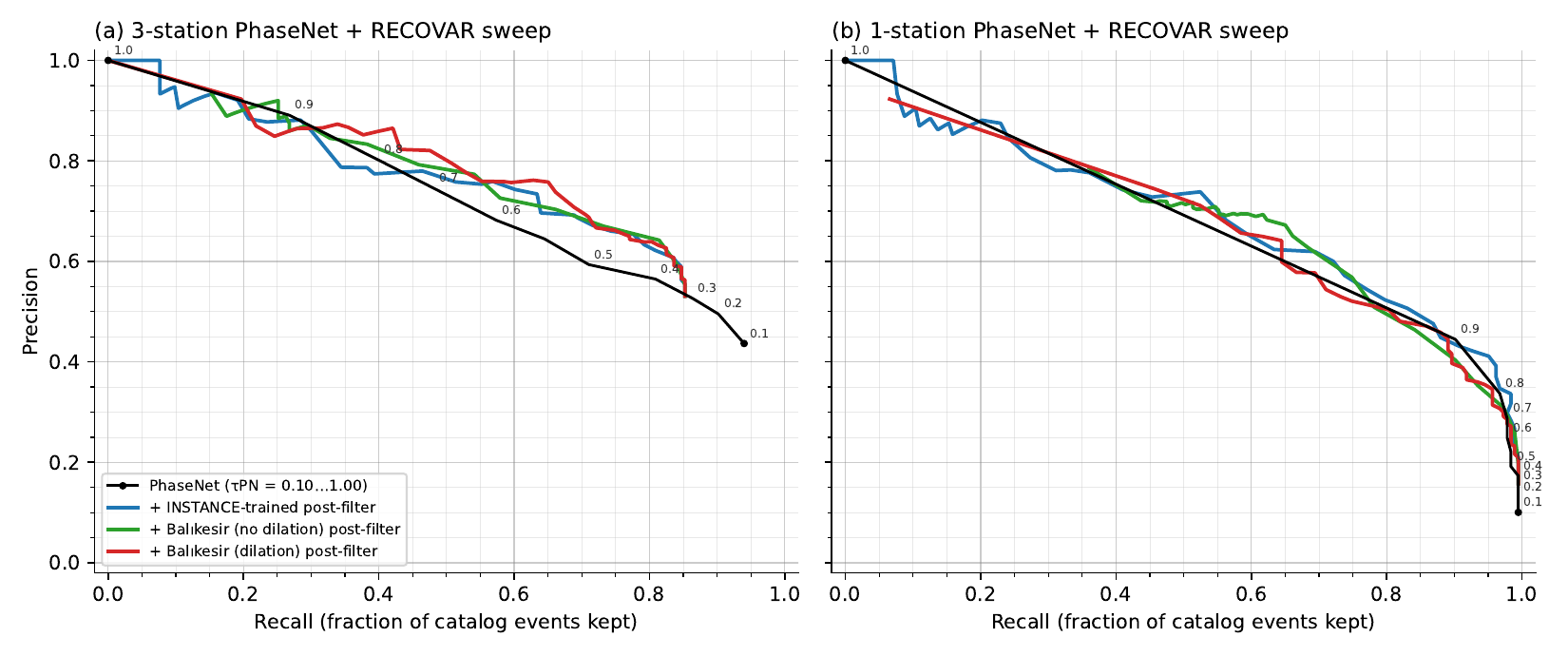}
    \caption{Precision versus recall for the Bal{\i}kesir--S{\i}nd{\i}rg{\i} evaluation set under (a) 3-station and (b) 1-station coincidence criteria. The black curve represents PhaseNet alone, with its threshold swept and the values $\tau_{PN} \in \{0.10, 0.20, \dots, 0.80, 1.0\}$ annotated. The colored curves represent the PhaseNet--RECOVAR detection-filtering pipelines, with PhaseNet fixed at $\tau_{PN}=0.30$ and the RECOVAR threshold $\tau_R$ swept. Recall is the fraction of the 183 cataloged events which are detected by the models. Under 1-station coincidence criterion, PhaseNet produces substantially more false detections and therefore operates at lower precision.}
\label{fig:pr_overlay}
\end{figure}

Figure~\ref{fig:pr_overlay} compares PhaseNet alone with the PhaseNet--RECOVAR hybrid pipelines, in which RECOVAR is used to filter PhaseNet detections. For PhaseNet alone, the onset-probability threshold is swept from 0.10 to 1.0. For the hybrid pipelines, the PhaseNet threshold is fixed at $\tau_{PN}=0.30$, while the RECOVAR threshold $\tau_R$ is swept. Because RECOVAR filters existing PhaseNet detections, the recall of each hybrid pipeline cannot exceed that of PhaseNet at $\tau_{PN}=0.30$. The corresponding recall ceilings are 0.852 under 3-station coincidence criterion, for which 156 of the 183 cataloged events are detected, and 0.995 under 1-station coincidence criterion, for which 182 events are detected.

\begin{table}[!ht]
\renewcommand{\arraystretch}{1.15}
\scriptsize
\setlength{\tabcolsep}{2.5pt}
\centering
\caption{Representative operating points for the PhaseNet--RECOVAR hybrid
pipelines on the Bal{\i}kesir--S{\i}nd{\i}rg{\i} evaluation set, with the
PhaseNet threshold fixed at $\tau_{PN}=0.30$. The RECOVAR threshold is
selected either to maximize false-positive filtering while preserving the
PhaseNet recall (\emph{best recall}) or to maximize the resulting $F_1$
score (\emph{best $F_1$}). The PhaseNet baseline is
$(\textrm{precision},\textrm{recall},F_1)=(0.531,0.852,0.654)$ under the
3-station coincidence criterion. The quantities $\Delta \textrm{Precision}$ and $\Delta \textrm{Recall}$
denote the absolute changes in precision and recall, respectively, relative
to the corresponding PhaseNet baseline. ``FP filtered'' and ``TP lost''
denote the percentage reductions in false-positive and true-positive
detections, respectively, relative to the PhaseNet baseline.}
\label{tab:postfilter_ops}
\begin{tabular*}{\textwidth}
{@{\extracolsep{\fill}}|l|cc|rrrr|}
\hline
&
\multicolumn{2}{c|}{\textbf{Best recall}} &
\multicolumn{4}{c|}{\textbf{Best \boldmath{$F_1$}}} \\
\cline{2-7}
\textbf{RECOVAR model}
& \boldmath{$\Delta \textrm{Precision}$}
& \textbf{\textrm{FP filtered}}
& \boldmath{$\Delta \textrm{Precision}$}
& \boldmath{$\Delta \textrm{Recall}$}
& \textbf{\textrm{TP lost}}
& \textbf{\textrm{FP filtered}} \\
\hline
INSTANCE-trained
& $+0.026$ & 10.1\%
& $+0.120$ & $-0.076$ & 9.0\% & \textbf{44.9\%} \\
Bal{\i}kesir--S{\i}nd{\i}rg{\i}, no dilation
& $\mathbf{+0.032}$ & \textbf{12.3\%}
& $\mathbf{+0.121}$ & $-0.043$ & 5.1\% & 42.8\% \\
Bal{\i}kesir--S{\i}nd{\i}rg{\i}, with dilation
& $+0.028$ & 10.9\%
& $+0.106$ & $\mathbf{-0.038}$ & \textbf{4.5\%} & 38.4\% \\
\hline
\end{tabular*}%

\end{table}

Across the recall range $[0.45,0.8]$, post-filtering with RECOVAR improves precision under both the 1-station and 3-station coincidence criteria. Under the 1-station criterion, the improvement is less consistent and marginal. Notably, the separation between the hybrid pipelines and PhaseNet is more significant under the 3-station criterion, implying that RECOVAR can eliminate spurious events which remain after multi-station coincidence criterion.

Table~\ref{tab:postfilter_ops} reports representative performance improvements of different RECOVAR variants under 3-station coincidence criterion for best recall and best $F_1$ threshold selection objectives. At $\tau_{PN}=0.30$, PhaseNet alone detects 156 catalog events and produces 138 false detections, yielding a recall of 0.852 and a precision of 0.531. Filtering by RECOVAR can reduce false detections improving the precision upto $0.032$ while preserving the same recall. At the best-recall operating point, all three models remove 10--12\% of the false detections without losing any of the 156 catalog-matched detections.

Allowing a small reduction in recall produces a larger precision gain. When operated at the threshold corresponding to the best-$F_1$ score\footnote{See Section~\ref{sec:evaluation-metrics} for the definition of $F_1$ score}, the hybrid pipelines remove 38--45\% of the false detections while keeping recall above 0.77, raising precision to 0.64--0.65 and $F_1$ to 0.71--0.72. The Bal{\i}kesir--S{\i}nd{\i}rg{\i}-trained model with dilation, for example, reaches $F_1=0.715$ at recall 0.814 and precision 0.637, compared with the PhaseNet baseline of $F_1=0.654$. More aggressive thresholds raise precision further at the cost of recall, as shown in Fig.~\ref{fig:pr_overlay}. Representative retained and rejected detections are shown in Fig.~\ref{fig:panels_grid}.

\section{Discussion and Conclusion}
\label{sec:conclusion}

We proposed RECOVAR, an unsupervised approach for seismic signal/noise classification. We trained and evaluated the method on public benchmark datasets and regional waveform data. On the public datasets, we followed the training and testing pipeline proposed by M\"unchmeyer \emph{et al.} \cite{rev:munchmeyer2022picker}. RECOVAR achieved performance comparable to that of supervised detectors while exhibiting relatively limited degradation under cross-dataset evaluation. This result suggests that the method is less sensitive to the choice of training dataset.

However, state-of-the-art supervised architectures can perform better under different training scenarios, and will most likely see further improvements in performance as dataset size and label quality increase over time. The value of unsupervised methods will be most apparent in conditions where such datasets are limited, rather than in direct competition with supervised approaches under benchmark conditions. Event-noise classification with waveforms from broadband seismometers is a domain which has seen a lot of high-quality work on data preparation, but as the landscape of seismological problems expand, new settings appear with limited access
to such high-quality, labeled data. Unsupervised methods can in principle be used in such settings, however, this requires the models to work with continuous waveforms rather than just benchmark datasets with removed labels.

To accommodate such a setting, we introduced a procedure that we refer to as batch dilation and used it to train the model on continuous regional waveform data. In these data, event-containing windows constituted less than 1\% of the training set, resulting in severe class imbalance; nevertheless, RECOVAR remained effective.

We also showed that RECOVAR can complement supervised detection pipelines. When used as a post-detection filter for PhaseNet detections, it improved precision in a multi-station coincidence setting. This capability to filter spurious event detections is relevant to commonly used event-localization pipelines.

Despite these advantages, RECOVAR focuses on waveform-level signal/noise classification rather than phase picking. The method must therefore currently be used together with a picking algorithm when arrival times or event locations are required. In future work, we plan to extend the latent-representation approach to unsupervised phase picking by using the temporal structure of the learned representations to estimate signal arrival times.

The current methodology is empirically motivated: seismic events induce shared latent structure across independently trained autoencoders, whereas noise produces weaker agreement. Future work could investigate the theoretical basis of this behavior and the conditions under which cross-representation covariance provides a reliable detection criterion. Because the detection principle makes no domain-specific assumptions, we also plan to test the method on signal-detection problems beyond seismology.

\section{Data Availability}
\label{sec:data}

The STEAD~\cite{mousavi2019stead} and INSTANCE~\cite{michelini2021instance} benchmark datasets are publicly available. The continuous waveforms for the Silivri and Bal{\i}kesir--S{\i}nd{\i}rg{\i} regions were retrieved through FDSN web services from the Kandilli Observatory and Earthquake Research Institute (KOERI) network (FDSN code KO,~\cite{koeri_network}). Station codes and time intervals are given in Section~\ref{sec:raw-datasets-and-training}. For labeling, the Silivri training catalog is from Durand \emph{et al.} \cite{durand2020twoscale}, and the Bal{\i}kesir--S{\i}nd{\i}rg{\i} per-station P arrival times were obtained from the online event bulletin of KOERI (\url{http://sc3.koeri.boun.edu.tr/eqevents/}, last accessed April 2026). The PhaseNet pick-filter evaluation catalog is from the European-Mediterranean Seismological Centre~\cite{emsc_catalog}. Source code and trained models are available at \url{https://github.com/onurefe/recovar} (last accessed April 2026).

\section*{Acknowledgment}
The authors would like to thank Dr. Yaman \"Ozak{\i}n, Dr. Hayrullah Karabulut, and Dr. Birsen Can for helpful discussions and guidance. They are also grateful for the contributions of former EarthML research group members in building the foundations of this study.

\section*{Conflicts of Interest}
The authors declare no conflicts of interest.

\bibliographystyle{unsrtnat}
\bibliography{biblio}

\end{document}